\documentclass[10pt,twocolumn]{article}

\usepackage{hyperref}
\usepackage{cleveref}
\usepackage{iccv}
\usepackage{times}
\usepackage{epsfig,enumitem}
\usepackage{graphicx}
\usepackage{amsmath}
\usepackage{amssymb}
\usepackage{mathtools, cuted}
\usepackage{bbm}
\usepackage{epstopdf}
\usepackage[font=footnotesize,caption=false]{subfig}
\usepackage{algorithm}
\usepackage[noend]{algpseudocode}
\usepackage{hhline}
\usepackage{multirow}
\usepackage{tabularx}
\usepackage{colortbl,cite}

\iccvfinalcopy 

\begin{document}

\title{An Integrated Approach to Crowd Video Analysis: From Tracking to Multi-level Activity Recognition}

\author{Neha Bhargava\\
Indian Institute of Technology Bombay\\
India\\
{\tt \small neha@ee.iitb.ac.in}
\and
Subhasis Chaudhuri\\
Indian Institute of Technology Bombay\\
India\\
{\tt\small sc@ee.iitb.ac.in}
}

\maketitle

\begin{abstract}
We present an integrated framework for simultaneous tracking, group detection and multi-level activity recognition in crowd videos. Instead of solving these problems independently and sequentially, we solve them together in a unified framework to utilize the strong correlation that exists among individual motion, groups and activities. We explore the hierarchical structure hidden in the video that connects individuals over time to produce tracks, connects individuals to form groups and also connects groups together to form a crowd. We show that estimation of this hidden structure corresponds to track association and group detection. We estimate this hidden structure under a linear programming formulation. The obtained graphical representation is further explored to recognize the node values that corresponds to multi-level activity recognition. This problem is solved under a structured SVM framework. The results on publicly available dataset show very competitive performance at all levels of granularity with the \textit{state-of-the-art} batch processing methods despite the proposed technique being an online (causal) one.
\end{abstract}

\setlength{\arrayrulewidth}{0.4mm}

\section{Introduction}
\begin{figure}[!h]
\centering
\subfloat{\includegraphics[scale=0.36]{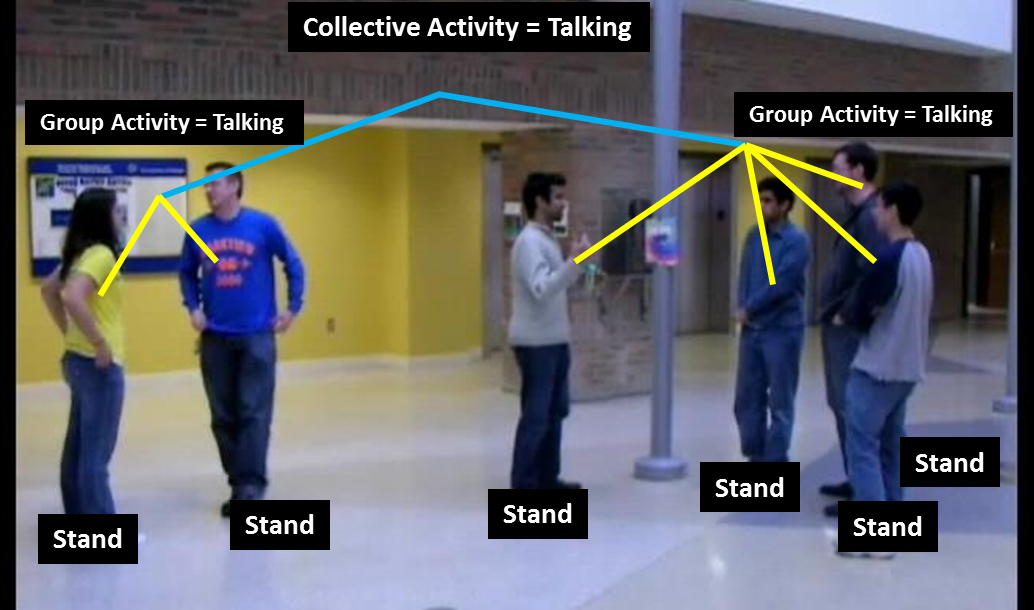}}
 \caption{Illustration of hierarchical structure present in a video. It represents video in terms of atomic actions, groups, group activities and collective activity. There are 6 individuals who are \textit{standing} and forming two groups with group activities as \textit{talking} and hence the collective activity is also \textit{talking}. }
\label{fig:intro_pic}
\end{figure}

A crowd video analysis system first detects the individuals and then tracks them over time. These tracks are used for higher level applications such as group detection and activity recognition. This approach is sequential in nature whereas the various components of the system are highly correlated and influence each other. For example, a particular group activity motivates its group member for a particular action and all the groups together define the crowd activity. On the other hand, group behavior is influenced by its members and the overall crowd behavior. Effectively, these components - individual's motion, groups, group activity and collective activity are correlated and can be expressed in a hierarchical structure. Hence it is more appropriate to estimate them together instead of sequentially. See Figure~\ref{fig:intro_pic} as an example of this hierarchical structure where the atomic actions of the individuals are all \textit{standing}, there are two groups each with group activity as \textit{talking} and thus leading to the collective activity also as \textit{talking}. These inherent dependencies among the various components motivate us to explore this idea of simultaneous recognition of tracks, groups and activities. We propose a novel approach to build on the detections to obtain the tracks, groups and activities in a causal framework, \ie without considering future frames into estimation process. We solve this unified problem in two passes. The first pass consists of finding the graph structure that corresponds to the track association and group detection. We propose an linear programming based formulation for the same. The second pass involves activity recognition at various levels of granularity. We formulate this problem under the structured SVM formulation \cite{ssvm}.

\begin{figure*}[htbp]
\centering
\subfloat[\label{fig:g2}]{\includegraphics[scale=0.35]{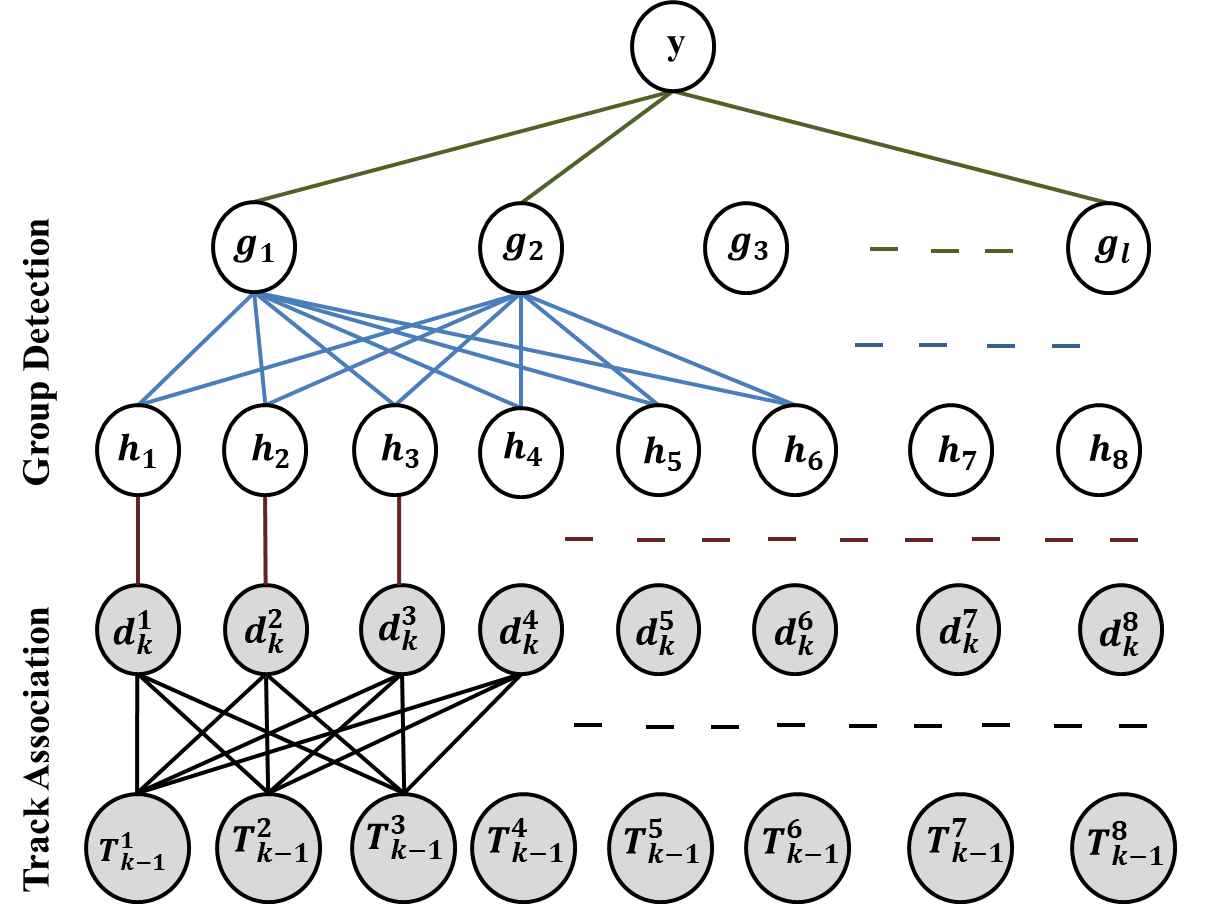}}\hspace{1cm}
\subfloat[\label{fig:g1}]{\includegraphics[scale=0.35]{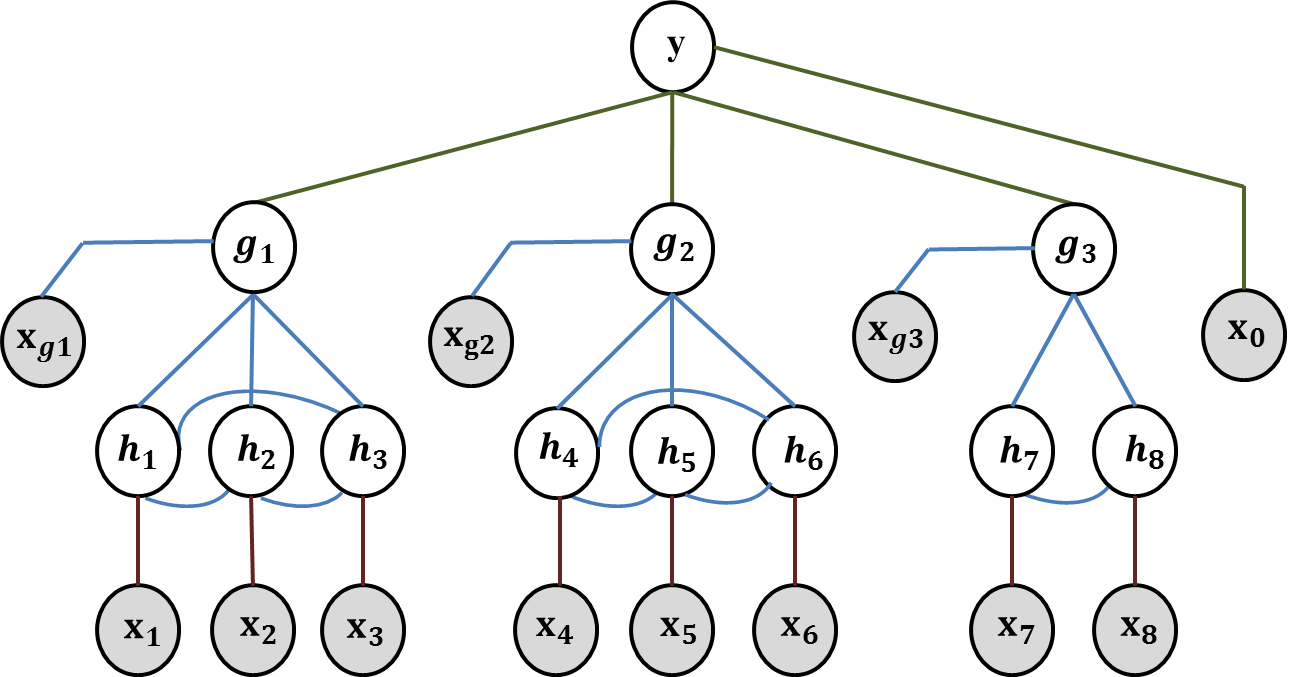}}
\caption{(a). Graphical representation showing the hierarchical structure present in a video. The first layer $\mathbf T_{k-1}$ and second layer $\mathbf d_k$ are fully connected since the track association is unknown until estimated. The third layer corresponds to the actions $\mathbf h_k$ associated with the detections. The third and fourth layers are again fully connected since the group association is unknown. The final layer corresponds to the connection between the overall activity and the group activities. (b). The figure shows a possible graphical structure obtained after track association and group detection. $\mathbf x_i$, $\mathbf x_g $ and $\mathbf x_0 $ are the respective features for a person, group and collectively that are derived from the video frames and to be used for activity recognition.}
\end{figure*}

In this paper, the term \textit{action} refers to an atomic movement performed by a single person, the term \textit{group activity} denotes an activity performed by a group and \textit{collective activity} refers to the activity performed by all the groups collectively. The paper is organized as follows. The next section discusses the related work followed by our contributions. The proposed model is described in Section \ref{section: model}. The subsequent Sections \ref{section: MTT_GT}, \ref{section:act_recog} and \ref{section:L_I} elaborate on frameworks for multi-target tracking with group detection followed by activity recognition. Experimentation details are provided in Section \ref{section:exp} and the paper concludes in Section~\ref{section:con}.

\section{Related work}
\label{section: lit}
The task of multi-target tracking (MTT) is to correctly associate all the detections (or tracklets) corresponding to each individual. Linear programming (LP) based global optimization for MTT is a popular approach. Many approaches formulate MTT either as min-cost flow optimization problem or MAP and use LP to find the global optimum. \cite{MTT_LP0, MTT_LP1, MTT_LP2, MTT_LP3, MTT_LP4}. Recently, the approaches of utilizing social behavior to improve tracking are gaining attention \cite{MTT_G0, MTT_G1, MTT_G2, MTT_G3, MTT_G4 }. The idea is to simultaneously associate a detection to a track and to a group. Our approach for obtaining groups is similar to that of \cite{MTT_G2} where they combine track association with grouping under a LP framework. Since the number of groups $K$ is unknown, they run the algorithm with all possible values of $K$ with a linear penalty. Our proposed method exploits the group information from the previous time instant and does not require to run for all values of $K$ resulting in a fast convergence. 

Due to its various applications in video surveillance, activity recognition has been an active area of research. The survey on action and activity recognition can be found in \cite{act_survey_0, act_survey_1, act_survey_2}. There are many works dealing with single person action \cite{action0, action1, action2, action3, action4} and with single group activity recognition \cite{choi2011, grp_act3, grp_act2, grp_act1, grp_act0}. Recently, researchers have started exploring the problem of joint recognition of these actions and activities under a hierarchical framework \cite{beyond, beyond_pami, choi2012, joint1, eccv2014}.  Amer \etal in \cite{eccv2014} proposed a hierarchical random field based modeling of higher order temporal dependencies  of video features. Lan \etal in \cite{beyond_pami} jointly estimate actions, pairwise interactions and group activity. However, they assume the availability of action labels and they do not handle track association. Choi and Savarese in \cite{choi2012} proposed a hierarchical model and combine the problems of tracklet association and multi-level activity recognition (action, pairwise interaction and collective). All these methods either assumed availability of \textit{action} label or \textit{trackelets} whereas our proposed framework requires only detections.

Our work in this paper advances the existing approaches and add one more intermediate layer (\ie \textit{grouping} layer) in the hierarchy as shown in Figure~\ref{fig:g2} and explained in Section~\ref{section: model}. The main contributions of this work are as follows:

\begin{enumerate}[noitemsep]
\item We propose a hierarchical graphical structure that combines multi-target tracking, group detection and activity recognition under an unified framework.
\item We built a \textit{causal} framework that takes only human detections as an input and outputs tracks, groups and activities at each time step.
\item We propose an iterative linear programming based method for simultaneous track association and group detection.
\item We propose an approach for simultaneous recognition of activities at various levels of granularity under a structured SVM framework.
\item To make it suitable for real-time applications, we provide a fast algorithm for both training and inferencing. 
\end{enumerate}

\section{The Proposed Model}
\label{section: model}
In this section, we discuss the proposed model. Let $y_k \in \mathcal Y$ be the collective activity at time $k$ with group activity vector $\mathbf g_k=[g_{1k}, g_{2k}, ..., g_{mk}]$ where $g_{ik} \in \mathcal G$ is the activity of $i^{th}$ group and $m$ be the number of groups present at time instant $k$. The atomic activity vector is denoted as $\mathbf h_k=[h_{1k}, h_{2k}, ..., h_{kN}]$ with $h_{ik} \in \mathcal H$ as the atomic activity of the $i^{th}$ person and $N$ is the total number of persons present at time $k$. Let $\mathbf T_{k-1}$ denotes the estimated tracks available till time $k-1$ and $\mathbf G_k$ be the group label vector of length $N$ where its $i^{th}$ entry denotes the group label for the $i^{th}$ detection at time $k$. Let $\mathbf d_k$ denotes the detections at time $k$. By a detection, we mean a person's location in the form of a bounding box. Now the problem statement is as follows: Given the detections $\mathbf d_k$ and tracks $\mathbf T_{k-1}$ at time $k$, the goals are (a) to associate these detections $\mathbf d_k$ to the appropriate tracks in $\mathbf T_{k-1}$ to get $\mathbf T_k$, (b) identify the group label vector $\mathbf G_k$ and (c) recognize the atomic, group and collective activities ($\mathbf h_k, \mathbf g_k$, $y_k$). Let $\mathbf z_k = [y_k,\mathbf g_k, \mathbf h_k]$ be the activity vector for notational convenience. The problem is formulated under the score maximization framework with a linear function as, 

\begin{equation}
\mathbf z_k^*,\mathbf G_k^*, \mathbf {T}_k^* = \arg\max \limits_{\mathbf z_k,\mathbf {G}_k,\mathbf T_k} \mathbf{w}^T {\Phi}(\mathbf z_k,\mathbf G_k,\mathbf T_k; \mathbf d_k,\mathbf T_{k-1}).
\end{equation}

The problem is illustrated as a graphical model in Figure \ref{fig:g2}. There are $N$ detections with an unknown number $N_g$ of groups at time $k$. $\mathbf T_{k-1}^i$ denotes the $i^{th}$ track available at time $k-1$. The root node denotes the collective activity which is connected to the group activity nodes. The group activity nodes are also connected to the atomic activity nodes of the group members. The graph emphasizes that collective activity is related to the group activities while a group activity is correlated both with its members' actions and the collective activity of the scene. Since the track association ($\mathbf T_{k-1} \leftrightarrow \mathbf d_k$) and group information ($h_i \leftrightarrow g_j$) are unknown - (a) every node $\mathbf T_{k-1}^i$ is connected to all the detection nodes and (b) each node of the action layer is connected to all the group activity nodes as shown in Figure~\ref{fig:g2}. Once we know the track association and group labels, the corresponding graph structure is known. One possible graph structure corresponding to Figure~\ref{fig:g2} is shown in Figure~\ref{fig:g1}. Here $\mathbf x_i$, $\mathbf x_g $ and $\mathbf x_0 $ are the respective observations for a person, group and collective entity defining the video. The procedure to obtain these observations are discussed later. 

We break this complete problem in two sub problems - (a) Graph structure estimation: This corresponds to track association and group detection (Eq.~\ref{eq:GT}), and (b) Node value estimation: This corresponds to multi-level activity recognition (Eq.~\ref{eq:A}). \ie

\begin{equation}
\mathbf {G}_k^*, \mathbf {T}_k^* = \arg\max \limits_{\mathbf {G}_k,\mathbf T_k} \mathbf{w_1}^T \Phi_1(\mathbf G_k,\mathbf T_k;\mathbf d_k,\mathbf T_{k-1},\mathbf z_{k-1})
\label{eq:GT}
\end{equation}
\begin{equation}
\mathbf z_{k}^* = \arg\max \limits_{\mathbf z_{k}}\mathbf{w_2}^T \Phi_2(\mathbf z_{k};\mathbf T_{k},\mathbf G_{k},\mathbf z_{k-1}).
\label{eq:A}
\end{equation}

The next two sections discuss these two sub problems in detail.

\section{Multi target tracking (MTT) and group detection}
\label{section: MTT_GT}
We estimate the tracks and groups together under a linear programming framework. Let $N$ number of detections and $N_g$ (unknown) number of groups be present at time $k$. Let $\mathbf T_{k-1}$ be a set of $T$ trajectories present at $k-1$. We define $\Psi \in \{0,1\}^{N\times T}$ as the track association matrix where $\Psi_{ij}=1$ indicates association of the $i^{th}$ detection with the $j^{th}$ track. We also define $\Omega \in \{0,1\}^{N\times N_g}$ as the group association matrix where $\Omega_{il}=1$ indicates that the $i^{th}$ detection belongs to the $l^{th}$ group. Then the optimization equation to estimate $\Psi$ and $\Omega$ is as follows:
\begin{equation}
\Psi^*, \Omega^* = \arg\max \limits_{\Psi, \Omega} \sum_{i=1}^{N} \sum_{j=1}^{T} \Psi_{ij}M_{ij}+\lambda\sum_{i=1}^{N} \sum_{l=1}^{N_g} \Omega_{il}C_{il}
\label{eq:TAG}
\end{equation}
\textit{s.t.}
\begin{eqnarray}
\Psi_{ij}, \Omega_{ij} \in \{0,1\},~
\sum_{j=1}^{N_g} \Omega_{il} \leq 1 ~\forall i, ~\nonumber \\
\sum_{i=1}^{N} \Psi_{ij} \leq 1 ~\forall j,
\sum_{j=1}^{T} \Psi_{ij} \leq 1 ~\forall i,~ 
\label{eq:constraints}
\end{eqnarray}
where $\lambda \in \mathbb R^+$ is a weighing factor that decides the balancing between the group association and track association scores. $M_{ij} \in [-1,1]$ is the compatibility score of the $i^{th}$ detection with the $j^{th}$ track based on motion and visual similarity, and $C_{il} \in [-1,1]$ is the compatibility score for the $i^{th}$ detection with the $l^{th}$ group based on motion, spatial and pose compatibility. The constraints in Eq.~\ref{eq:constraints} ensure that each detection is assigned to at most one track and to one group. It also ensures that each track gets at most one detection while there is no such constraint for the group. The next sub-sections discuss the construction of compatibility matrices $M$ and $C$.


\subsection{Construction of $M$}
$M \in \mathbb R ^{N\times T}$ is a score matrix for track association where $M_{ij}$ is the score of assigning $i^{th}$ track to the $j^{th}$ detection. It is calculated based on visual similarity, spatial proximity and the motion compatibility between the $i^{th}$ detection with the $j^{th}$ track. 

The visual similarity score is based on color histogram matching of the $j^{th}$ detection with that at the last location of the $i^{th}$ track. The spatial proximity is the measure of closeness of the $j^{th}$ detection from the $i^{th}$ track. Lastly, the motion compatibility is based on the velocity consistency when the $j^{th}$ detection is added to the $i^{th}$ track. By combining these three scores, we obtain
\begin{equation}
M_{ij} = \sum_{n=1}^{3}\alpha_n (2e^{-\beta_n||\mathbf x^{(n)}_i-\mathbf x^{(n)}_j||^2_2}-1),
\end{equation}
where $\alpha$ and $\beta$ are the weight and normalizing vectors respectively. $\mathbf x^{(1)}$ represents color histogram, $\mathbf x^{(2)}$ is the location and $\mathbf x^{(3)}$ is the velocity estimate. We keep $\alpha_n=\frac{1}{3}$, $\beta_1=1$ and $\beta_3=1$ in the experiments. $\beta_2$ is chosen as the inverse of height of the bounding box.

\subsection{Construction of $C$}
\begin{figure}[!h]
\centering
\subfloat{\includegraphics[scale=0.35]{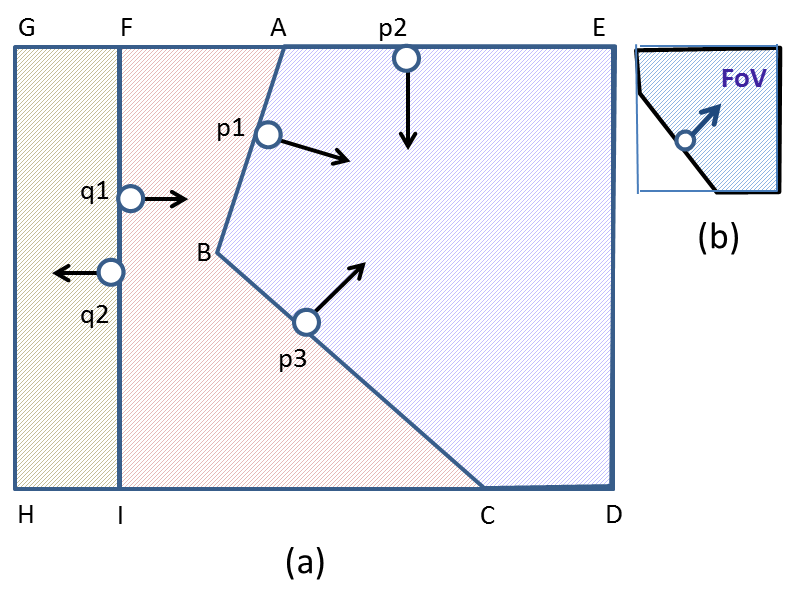}}
\caption{(a) Illustration of score calculation for pose compatibility between a candidate group and a detection. In the figure, (p1, p2, p3) form a group with group's FoV as $ABCDE$. $FIDE$ and $GHIF$ are FoVs of detections q1 and q2 respectively. Therefore $S(q1,(p1,p2,p3))=\frac{ABCDE \cap FIDE}{ABCDE}=1$ while $S(q2,(p1,p2,p3))=\frac{ABCDE \cap GHIF}{ABCDE}=0$. (b) Illustration of field of view (FoV) for a person. The arrow signifies the pose direction. The boundary rectangle corresponds to the observed image. Best viewd in color.}
\label{fig:pose_score}
\end{figure}

$C \in \mathbb R ^{N\times N_g}$ is a score matrix for group association where $C_{il}$ is the score of assigning the $i^{th}$ detection to the $l^{th}$ group. It is calculated based on the motion similarity, spatial closeness and the pose compatibility between the $i^{th}$ detection and the $l^{th}$ group. The group location and group velocity are defined as the averages over the locations and velocities of the members, respectively. To compute motion similarity between $i^{th}$ detection and $l^{th}$ group, we first find the track associated with the $i^{th}$ detection from $\Psi$. We then compute the velocity compatibility between the obtained track with the $l^{th}$ group. To obtain pose compatibility, we first calculate the interacting zone of the group formed by the members. The normalized intersection of the field of vision of the detection with the group's interacting zone gives the score for the pose compatibility. This is illustrated in Figure~\ref{fig:pose_score}. Let p1, p2 and p3 form a group and q1, q2 are the detections. We define field of view (FoV) for a person as the complete area in the pose direction as illustrated in Figure~\ref{fig:pose_score}(b). The pose compatibility between a detection $d$ and a group $\bar g$ is defined as $S(d, \bar g) = \frac{FoV(\bar g) \cap FoV(d)}{ FoV(\bar g)}$, where $FoV(\bar g)$ is the intersection of FoVs of the group members. In Figure~\ref{fig:pose_score}, q1 has high compatibility score while q2 has zero score since it has no intersection. The pose compatibility is added to discourage the non-facing persons forming a group. Finally, we combine the three scores obtained from motion, spatial and pose compatibilities to construct $C$ as done previously for $M$.

\subsection{Iterative algorithm to obtain $\Psi$ and $\Omega$}
Since the group information is initially unknown at time $k$, we do not know the score matrix for group association \ie $C$. Hence, we propose an iterative algorithm to construct $C$ and to solve Eq. \ref{eq:TAG}. We use the group information from the previous time instant $k-1$ to get an initial estimate of $C$ for the present detections. Then we solve Eq. \ref{eq:TAG} to get the optimal $\Omega$. If any row (say $i^{th}$) of $\Omega$ consists of all zeros, it indicates that $i^{th}$ detection does not belong to any of the groups. In such a case, we add one more group to the list with $i^{th}$ detection as its founding member and again solve for Eq. \ref{eq:TAG}. We iteratively do this until we get group assignment for all the detections. Also, to discourage formation of singleton groups (with one member) , we remove such groups before the start of the iterative algorithm at each frame. To initialize in the first frame of the video, we consider $N_g=1$ \ie all the detections belong to a single group. This iterative method is detailed in Algorithm~\ref{alg:PO}. The algorithm is found to converge within a few iterations only. In the worst case when all the detections form singleton groups and different from the groups present at previous time instant, the algorithm takes $N$ number of iterations.    

\begin{algorithm}
\caption{Algorithm to obtain $\Psi$ and $\Omega$}\label{alg:PO}
\begin{algorithmic}[1]
\Procedure{}{}
\State t=0, $\mathbf G^t_k = \mathbf G_{k-1}$ 
\State Solve Eq.~\ref{eq:TAG} to get $\Psi^*$ and $\Omega^*$
\State \textit{$\mathbf d$} = set of detections without any group assignment
\While{\textit{$\mathbf d$} is non-empty}
\begin{itemize}[noitemsep,nolistsep]
\item Add one column to $\Omega^*$ with one of the detections from \textit{$\mathbf d$}
\item Solve Eq.~\ref{eq:TAG} to get $\Psi^*$ and $\Omega^*$
\item Update $\mathbf G^t_k$ and \textit{id}
\item $t \gets t+1$
\end{itemize}
\EndWhile
\State \textbf{return} $\mathbf G^t_k$, $\Psi^*$ and $\Omega^*$
\EndProcedure
\textbf{end}
\end{algorithmic}
\end{algorithm}

\section{Activity recognition}
\label{section:act_recog}
Solution of Eq.~\ref{eq:TAG} gives an estimate of the latent graph structure (\eg Figure~\ref{fig:g1}). The next problem is to estimate the optimal node values of this graph structure at all time instants causally. In other words, the aim is to recognize the activities at individual, group and collective levels.

The problem is cast under a linear energy function framework as
\begin{eqnarray}
\Phi(y,\mathbf g, \mathbf h,\mathbf x) = \mathbf w^T \mathbf{\phi}(y,\mathbf g, \mathbf h,\mathbf x),
\label{eq:act_rec}
\end{eqnarray} 
where $\mathbf {\phi}$ calculates the compatibility of activities ($y, \mathbf h, \mathbf g$) and the observations $\mathbf x =\{\mathbf x_0, \mathbf x_g, \mathbf x_i\}$. We follow the motivation of  \cite{choi2012} to solve the problem. As said before, $\mathbf x$ contains individual, group and collective features which are obtained once $\mathbf T_k$ and $\mathbf G_k$ are known. We take advantage of hierarchical structure and decompose $\mathbf {\Phi}(y,\mathbf g,\mathbf h,\mathbf x)$ according to the graph Figure~\ref{fig:g1} as follows: 

\begin{eqnarray}
&\Phi(y,\mathbf g,\mathbf h,\mathbf x)=\textit w_0^T\phi_0(y,\mathbf x_0)+\textit w_1^T\phi_1(y,H(\mathbf g))\nonumber \\
&+\sum\limits_{i=1}^{N_g}\textit w_2^T\phi_2(g_i,x_{g_i})+\sum\limits_{i=1}^{N_g}\textit w_3^T\phi_3(g_i, H(\mathbf h_{g^i}))\nonumber \\
&+\textit w_4^T\sum\limits_{i=1}^{N}\phi_4(h_i, x_i),
\label{eq:decomp}
\end{eqnarray}

where {$\textbf{w} = [\textit{w}_0, \textit{w}_1, \textit{w}_2, \textit{w}_3, \textit{w}_4]$} and {$\phi=[\phi_0, \phi_1, \phi_2, \phi_3, \phi_4]$}. Each term is described as follows:
\begin{enumerate}
\item \textbf{Collective Activity - Image Potential}:\\
 It is the compatibility score of collective activity $y \in \mathcal Y$ with the collective observation $\mathbf x_0$. It is modeled as
\begin{equation}
\textit w_0^T\phi_0(y,\mathbf x_0)=\sum_{a \in \mathcal{Y}} \textit w_{0a}^T\mathbbm{1}(y=a)\mathbf x_0,
\end{equation}
where $\textit{w}_0 = [\textit{w}_{01}, \textit{w}_{02}, ..., \textit{w}_{0|\mathcal Y|}]$ and $\mathbbm{1}(:)$ is an indicator function.

\item \textbf{Collective - Group Activity Potential}: \\
$\textit w_1^T\phi_1(y,g)$ is the compatibility of group activities $\mathbf g$ with the collective activity $y$ and defined as
\begin{equation}
\textit w_1^T\phi_1(y,H(\mathbf g))=\sum_{a \in \mathcal{Y}} \textit w_{1a}^T\mathbbm{1}(y=a)H(\mathbf g),
\end{equation}
where $H(\mathbf g)$ is the histogram of group activities.
  
\item \textbf{Group Activity - Image Potential}: \\
$\textit w_2^T\phi_2(g,\mathbf x_{g})$ defines the compatibility of group activity $g \in \mathcal G$ with the group observation $\mathbf x_g$ as
\begin{equation}
\textit w_2^T\phi_2(g,\mathbf x_g)=\sum_{b \in \mathcal{G}} \textit w_{2b}^T\mathbbm{1}(g=b)\mathbf x_g.
\end{equation}

\item \textbf{Group Activity Potential}:\\
$\textit w_3^T\phi_3(g, H(h_g))$ defines the compatibility of atomic activities of the group members with the group activity. It is modeled as
\begin{equation}
\textit w_3^T\phi_2(g,H(h_g))=\sum_{b \in \mathcal{G}} \textit w_{3b}^T\mathbbm{1}(g=b)H(\mathbf h_g),
\end{equation}
where $H(\mathbf h_g)$ is the histogram of atomic activities of the group members.
 
\item \textbf{Atomic Action - Image Potential}:\\
$\textit w_4^T\phi_4(h_i,\mathbf x_i)$ defines the compatibility of the individual's observation with the atomic activity and modeled as
\begin{equation}
\textit w_4^T\phi_4(h_i,\mathbf x_i)=\sum_{c \in \mathcal{H}} \textit w_{4c}^T\mathbbm{1}(h_i=c)\mathbf x_i.
\end{equation}

\end{enumerate}

\begin{algorithm*}
\caption{Inference algorithm}\label{alg:IA}
\begin{algorithmic}[1]
\Procedure{Inference}{}
\State Initialize $y^0$, $\mathbf g^0$, $\mathbf h^0$
\State t=0, \textit{err}=1000 , $\epsilon =0.01$
\While{$\textit{err}>\epsilon$}
\State ${
y^{t+1} \gets \arg\max \limits_y \{\textit{w}_0^T\phi_0(y,\mathbf x_0)+\textit{w}_1^ T\phi_1(y,H(\mathbf g^{t}))}\}$
\State $g_i^{t+1} \gets \arg\max \limits_g\{\textit{w}_1^T\phi_1(y^{t+1},H(\mathbf g^t\backslash g_i^t,g))+\textit{w}_2^ T\phi_2(g_i,\mathbf x_g)+\textit{w}_3^T\phi_3(g,H(\mathbf h_g^t)\}$, $\forall i=1:N_g$
\State ${h_i^{t+1} \gets \arg\max \limits_h \{\textit{w}_3^T\phi_3(g_j^{t+1},H(\mathbf h^t \backslash h_i^t,h))+\textit{w}_4^ T\phi_4(\mathbf x_i,h)}\}$, $\forall i=1:N$ , {$j$: group index of $i^{th}$ person}
\State $\textit{err} \gets \frac{1}{1+N+N_g}\{{\mathbbm{1}(y^{t}\neq y^{t+1})+\mathbbm{1}(\mathbf g^{t}\neq \mathbf g^{t+1})+\mathbbm{1}(\mathbf h^{t}\neq \mathbf h^{t+1})}\}$
\State $t \gets t+1$
\EndWhile
\State \textbf{return} $y^t$, $\mathbf g^t$ and $\mathbf h^t$
\EndProcedure
\textbf{end}
\end{algorithmic}
\end{algorithm*}

\section{Inference and Learning for activity recognition}
\label{section:L_I}
In this section, we discuss the learning and inference algorithms. Given a graph structure at any time instant $k$ (\textit{i.e} $\mathbf T_k$ and $\mathbf G_k$), we need to recognize the activities at all levels. Solution of Eq.~\ref{eq:act_rec} provides the inference about the unknown node variables $(y, \mathbf g, \mathbf h)$. We use the structured SVM framework \cite{trssvm} to learn $\mathbf w$, and an iterative alternate optimization method for the inference. The next two subsections discuss both these algorithms in detail. 

\subsection{Inference}
Given the learned model parameters \textbf{w}, the inference problem is to find the optimal collective activity $y^*$, group activity vector $\mathbf g^*$ and atomic activity vector $\mathbf h^*$ for the input $\mathbf x$. \ie
\begin{eqnarray}
y^*, \mathbf g^*, \mathbf h^* = \arg \max \limits_{y, \mathbf g, \mathbf h} \mathbf w^T\phi(y,\mathbf g, \mathbf h, \mathbf x).
\label{eq:Inf}
\end{eqnarray}

We use an iterative method to solve Eq.~\ref{eq:Inf}. We initialize $y$, $\mathbf g$ and $\mathbf h$ with the values in the previous time step if available or random otherwise. The method is detailed in Algorithm~\ref{alg:IA}. 

\subsection{Learning}
Given a training data $D=\{\mathbf x^i, \mathbf G^i, \mathbf h^i, \mathbf g^i, y^i\}~\forall \{i=1,2, ..., S\}$ where $S$ is the total number of training samples and $\mathbf G^i$ is the group label vector, the goal is to learn the optimal weight vector $\textbf w^*$. We use 1-Slack structured SVM with margin-rescaling \cite{trssvm} where there is only a single slack variable $\xi$ for all the constraints. Let us define $\mathbf z^i = [\mathbf h^i, \mathbf g^i, y^i]$ to simplify the notations. The optimization equation is as follows:

\begin{eqnarray}
&\textbf w^* = \arg \min \limits_{\textbf w} \frac{1}{2}||\textbf{w}||_2^2 +D\xi
\end{eqnarray}
\textit{s.t.} $\forall \mathbf z^i:$
\begin{eqnarray}
\frac{1}{S}\textbf w^T\sum\limits_{i=1}^{S}[\phi(\mathbf x^i, \mathbf z^i)-\phi(\mathbf x^i, \bar{\textbf z}^i)]\geq\frac{1}{S}\sum\limits_{i=1}^{S}\Delta(\mathbf z^i, \bar{\textbf z}^i) -\xi.
\end{eqnarray}
The loss function $\Delta(\bar{\textbf z}, \mathbf z^i)$ is defined as 
\begin{eqnarray}
\Delta(\bar{\textbf z}, \mathbf z^i)=\frac{1}{|\mathbf z|}\sum\limits_{j=1}^{|\mathbf z|} \mathbbm{1} (\bar z_j \neq z^i_j),
\end{eqnarray}
where $\bar{\textbf z}$ is any possible combination and $\mathbf z^i$ is the actual output corresponding to the $i^{th}$ input.

Since the number of constraints grows exponentially with $S$, the cutting plane algorithm \cite{trssvm} constructs a set of working constraints and optimize the function over this set. This set is constructed by identifying the most violated constraint for each data sample $(\mathbf x^i, \mathbf z^i)$ at each iteration. Finding the most violated constraint for $(\mathbf x^i, \mathbf z^i)$ is again an optimization problem and is as follows:
\begin{equation}
\hat{\textbf z}^i = \arg\max\limits_{\mathbf z \in \mathbb Z} \mathbf w^T\phi(\mathbf z, \mathbf x^i)+\Delta(\mathbf z, \mathbf z^i).
\end{equation}

This is same as our inference problem with an additional term of $\Delta(\mathbf z, \mathbf z^i)$. We use the same method to solve this.

\section{Discussions and Experiments}
\label{section:exp}
\subsection{Dataset}
We demonstrate the performance of the proposed method on the commonly used collective activity dataset provided in \cite{choi2012}. The dataset has 44 video clips composed of different challenging videos. The annotations for 5 collective activities (\textit{crossing}, \textit{waiting}, \textit{queuing}, \textit{walking}, and \textit{talking}) and 8 poses (\textit{right}, \textit{right-front}, ..., \textit{etc.}) are provided. Additionally, the authors of \cite{choi2012} have provided annotations for target correspondence, atomic action labels (\textit{standing}, \textit{walking}) and 8 pairwise interaction labels. Since we are interested in finding groups and group activities instead of pair-wise interactions, we provide annotations for group labels and group activities (\textit{walking}, \textit{waiting}, \textit{queuing} and \textit{talking}) after every 10 frames. We consider collective activity as the major activity happening at a time. For example - if out of 5 groups, 3 or 4 groups are \textit{talking} and one is \textit{walking}, we consider the overall activity as \textit{talking}. Moreover, we differ in the definition of \textit{crossing} from that mentioned in \cite{choi2012}. In this paper, we consider \textit{crossing} happens when two or more groups cross each other on the contrary to road crossing used in \cite{choi2012}. We have re-annotated \textit{crossing} videos accordingly. 

As is common in most feature tracking methods, we pre-process the videos for image stabilization. To do this, we use a time window of 20 frames where the $1^{st}$ frame acts as the reference frame. The camera motion is compensated in the subsequent frames with respect to it by estimating an affine transformation between the reference frame and the $k^{th}$ frame. 

\subsection{Observations}
\begin{figure}[!h]
\centering
\subfloat[\textit{Waiting}]{\includegraphics[scale=0.5]{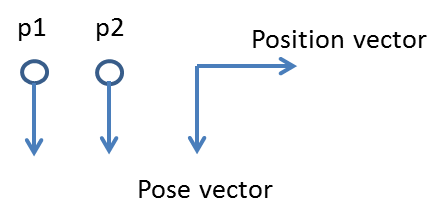}}
\subfloat[\textit{Queue}]{\includegraphics[scale=0.5]{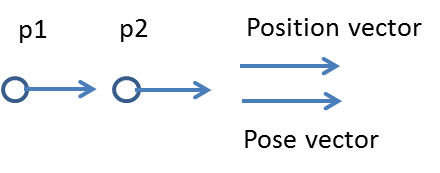}}
\caption{Illustration of pose-position compatibility score. The arrows for p1 and p2 indicate their pose directions. Basic setup in case of \textit{waiting} and \textit{queue} to be utilized to discriminate between them. (a) In case of \textit{waiting}, the persons p1 and p2 are standing side by side, thereby creating a right angle between position vector (p1-p2) with pose vector. (b) In case of \textit{queue}, the persons p1 and p2 are one after another and hence the position vector is aligned with pose vector.}
\label{fig:QW}
\end{figure}

The observations $\mathbf x$ consist of individual related features $\mathbf x_i$, group level features $\mathbf x_g$ and collective features $\mathbf x_0$. The individual observations $\mathbf x_i \in \mathbb R^{|\mathcal P|\times|\mathcal H|}$ include pose $\in \mathcal P$ and action $\in \mathcal H$. $\mathbf x_g$  is the mean of the feature vectors of the  group members while $\mathbf x_0$ is the mean of feature vectors of all the individuals. Note that only pose and action are not enough to discriminate between \textit{waiting} and \textit{queue} since all the members possess the same pose and action. To incorporate some discrimination, we additionally include a pose-position compatibility to $\mathbf x_g$. The score is calculated for all the pairs $(i, j)$ of the group members and is defined as $|p^T(d_i-d_j)|$ where $p$ is the pose vector corresponding to the statistical mode of the member poses and $d_i$ is the position of the $i^{th}$ member. Higher value of the score corresponds to \textit {queue} since both the vectors are aligned in the same direction while \textit{waiting} will have a low value because both the vectors are orthogonal to each other. This is illustrated in Figure~\ref{fig:QW}. We append the mean value of the score values obtained for all the pairs of the group to $\mathbf x_g$. 

To learn a pose classifier, we fine-tune all the 19 layers of the VGG \cite{Simonyan14c} network on PARSE-27k \cite{PARSE27k} pedestrian attribute dataset comprised of 27 thousand labeled training images. To account for inherent order in poses, we modify cross entropy loss by penalizing misclassification. The penalty is less for predicting nearby pose and high otherwise; For example, the penalty is less if the classifier predicts \textit{Right-Front} for the true pose of \textit{Right} while the penalty is high if the prediction is \textit{Left}. 

We employ the following procedure to estimate action. We fit lines separately on the $\textit{x}$ and $\textit{y}$ coordinates of the \textit{top-left} and \textit{bottom right} of the bounding box as a function of time over 20 frames and use the estimated slopes to learn a SVM classifier for atomic action classification. The reason for considering both \textit{top-left} and \textit{bottom right} coordinates of the bounding box is to capture the possible movement along the viewing direction of the camera (\ie effect of approaching and receding). 

\subsection{Tracking performance}
We assume that the detections per frame are available to us. We do not handle occlusion in this paper. Whenever any target returns back to the scene after occlusion, a new id is assigned to it. To evaluate the tracking performance, we consider the number of identity switches. We compare the tracking results with a baseline model present in our framework. It corresponds to the track association based on visual, spatial and velocity compatibility (first part of Eq. \ref{eq:TAG}) only. The full model incorporates both track association and group association. The number of ID switches are given in Table~\ref{table:trk_metrics}. The total number of tracks in the dataset is 466. The decrease in the number of ID switches in the full model indicates the effectiveness of combined estimation of groups and tracks over independent track association. 

\begin{table}[h!]
\small
\caption{Table showing tracking performance}
\centering
\begin{tabular}{ |c|c|c| } 
\hline
  & Baseline model & Full model \\
\hline 
ID Switches  & 22 (4.5\%) & 17 (3.7\%) \\ 
\hline
\end{tabular}
\label{table:trk_metrics}
\end{table}

\subsection{Group detection performance}
To evaluate the group detection performance, we use the following clustering measures which are commonly used: Purity \cite{purity}, Rand Index \cite{ri} and Normalized mutual information (NMI) \cite{nmi}. We compare with a baseline case present within our framework which corresponds to group association (second part of Eq. \ref{eq:TAG}). The full model incorporates both track association and group association. The quantitative results are given in Table \ref{table:grp_metrics}.
\begin{table}[h!]
\small
\caption{Table showing group detection performance}
\centering
\begin{tabular}{ |c|c|c|c| } 
\hline
Framework & Purity & Rand Index & NMI \\
\hline 
 Baseline & 0.82 & 0.75 & 0.65\\ 
 Full & 0.89 & 0.81 & 0.72\\ 
 \hline
\end{tabular}
\label{table:grp_metrics}
\end{table}

Again, the higher values of the clustering measures in the full model indicates the effectiveness of combined estimation of groups and tracks over independent group detection.

\subsection{Activity recognition performance}
\begin{figure*}[htbp]
\subfloat[]{\includegraphics[scale=0.3]{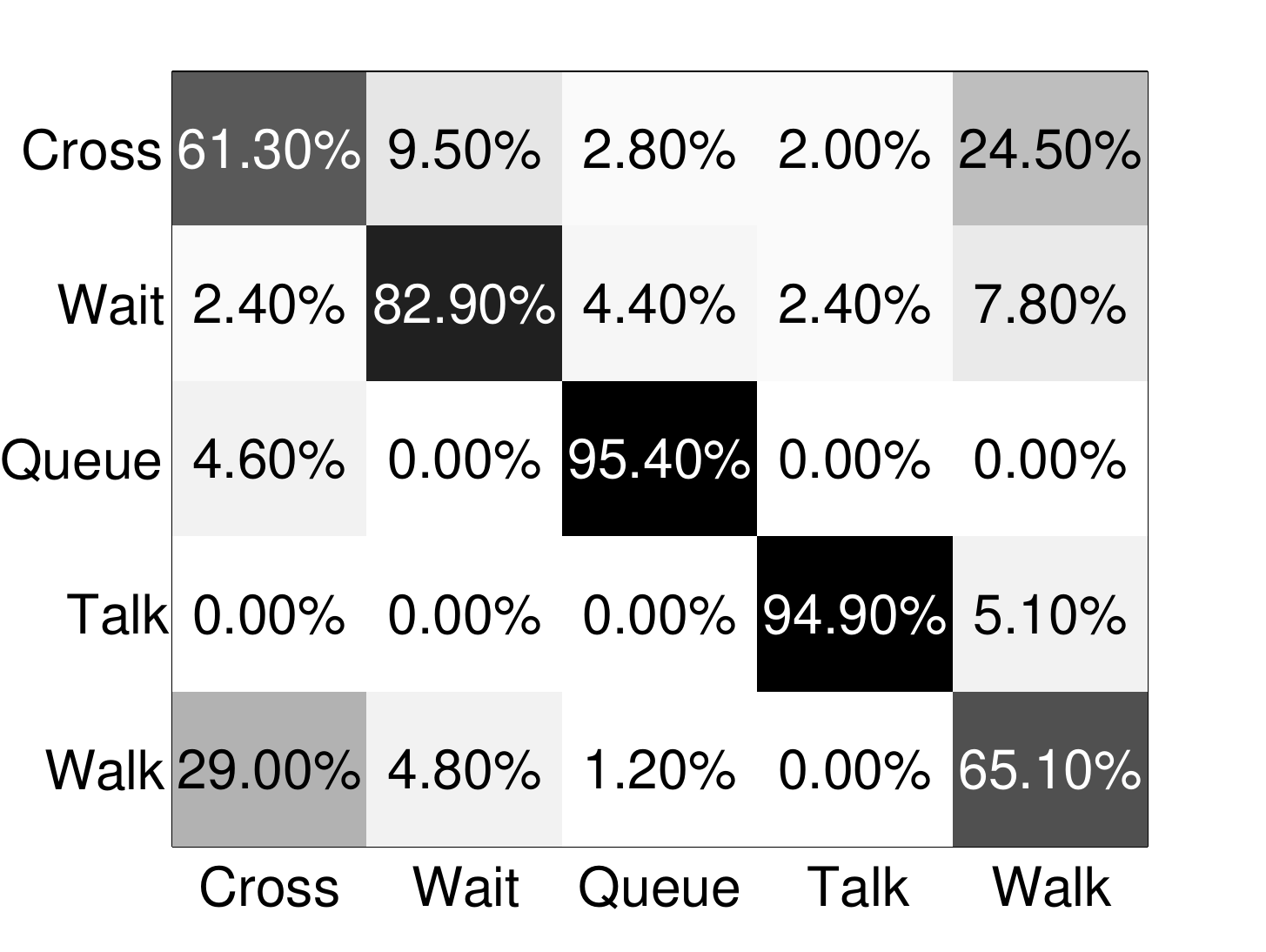}}
\subfloat[]{\includegraphics[scale=0.3]{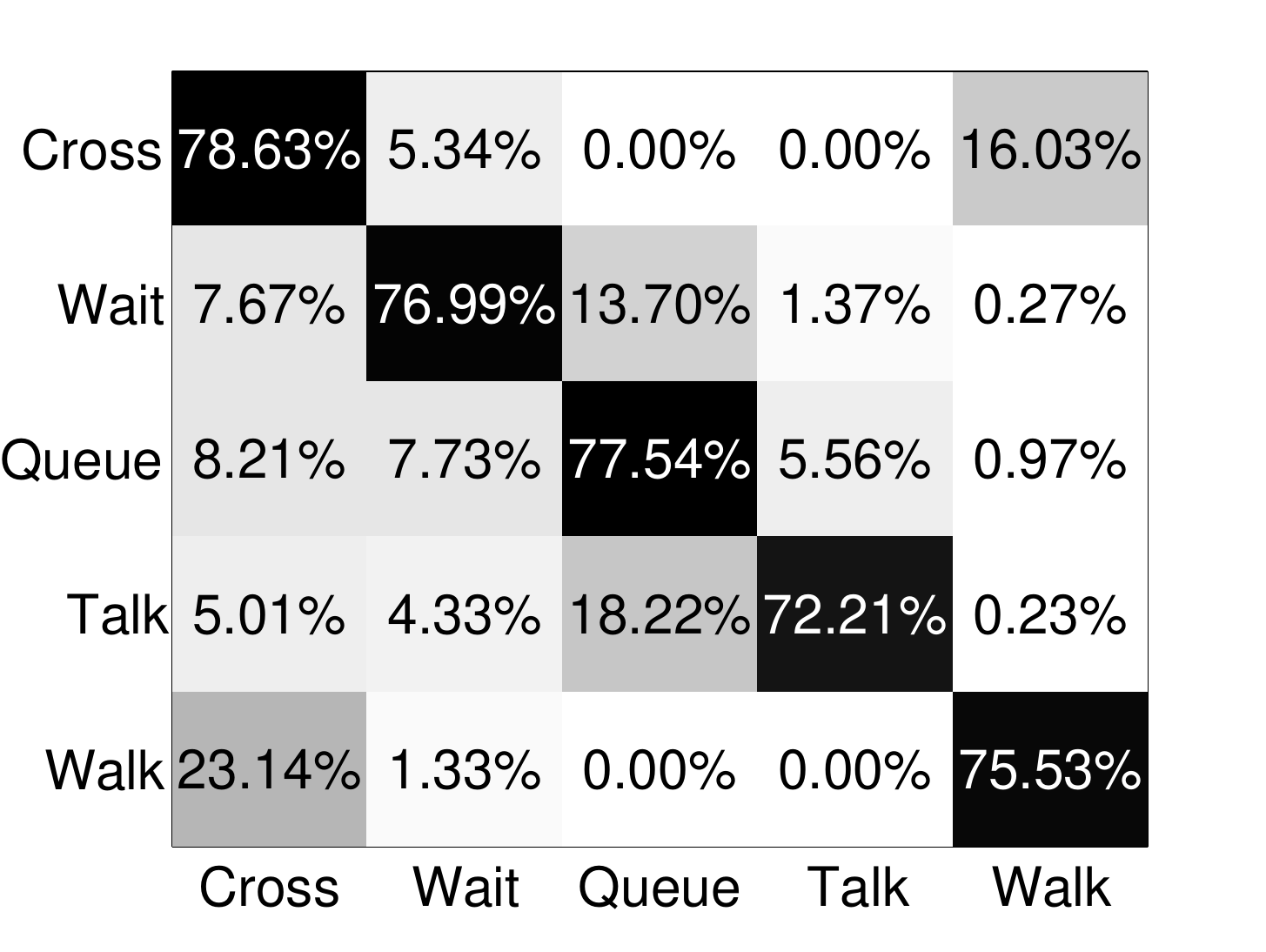}}
\subfloat[]{\includegraphics[scale=0.3]{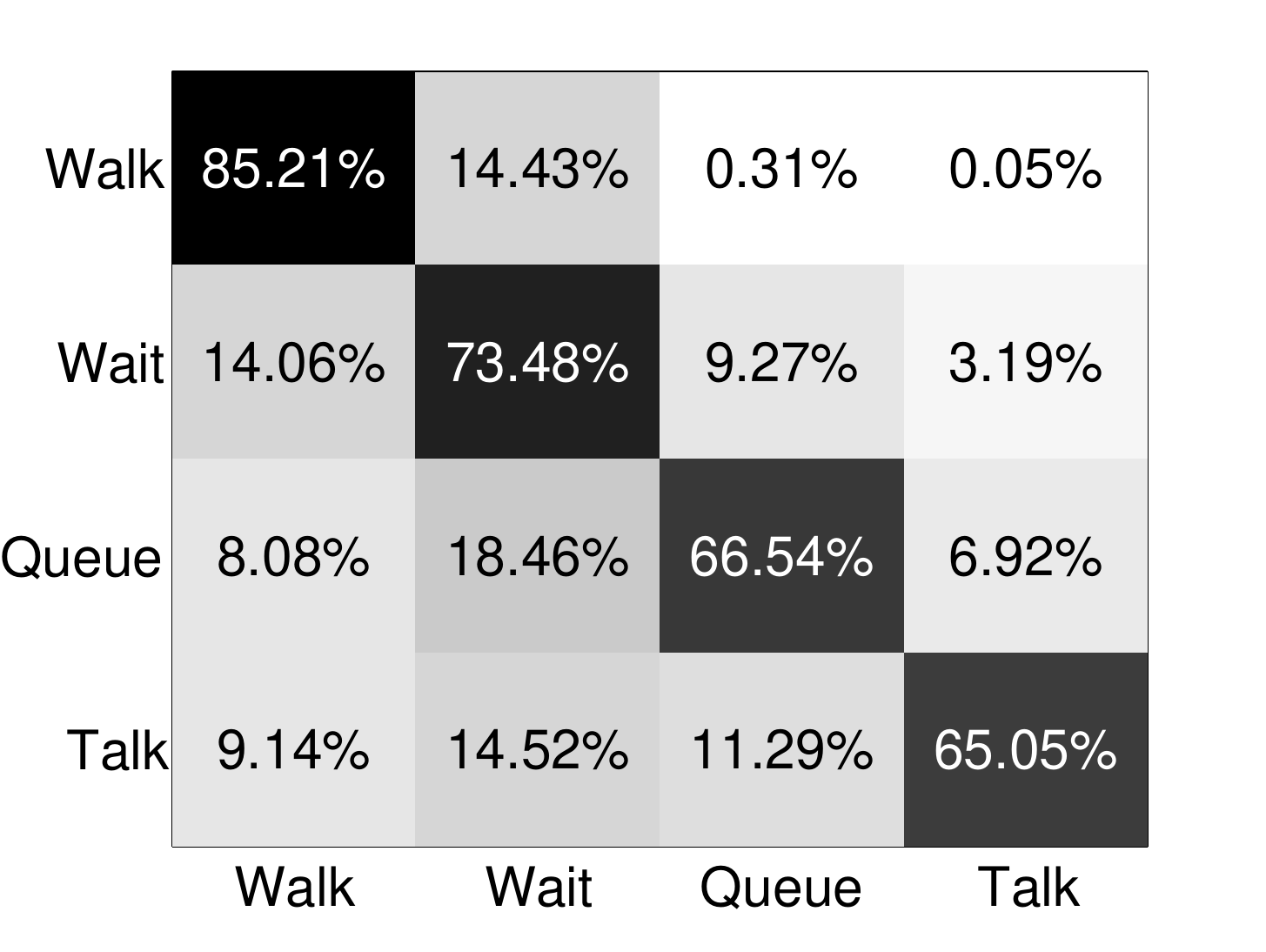}}
\subfloat[]{\includegraphics[scale=0.3]{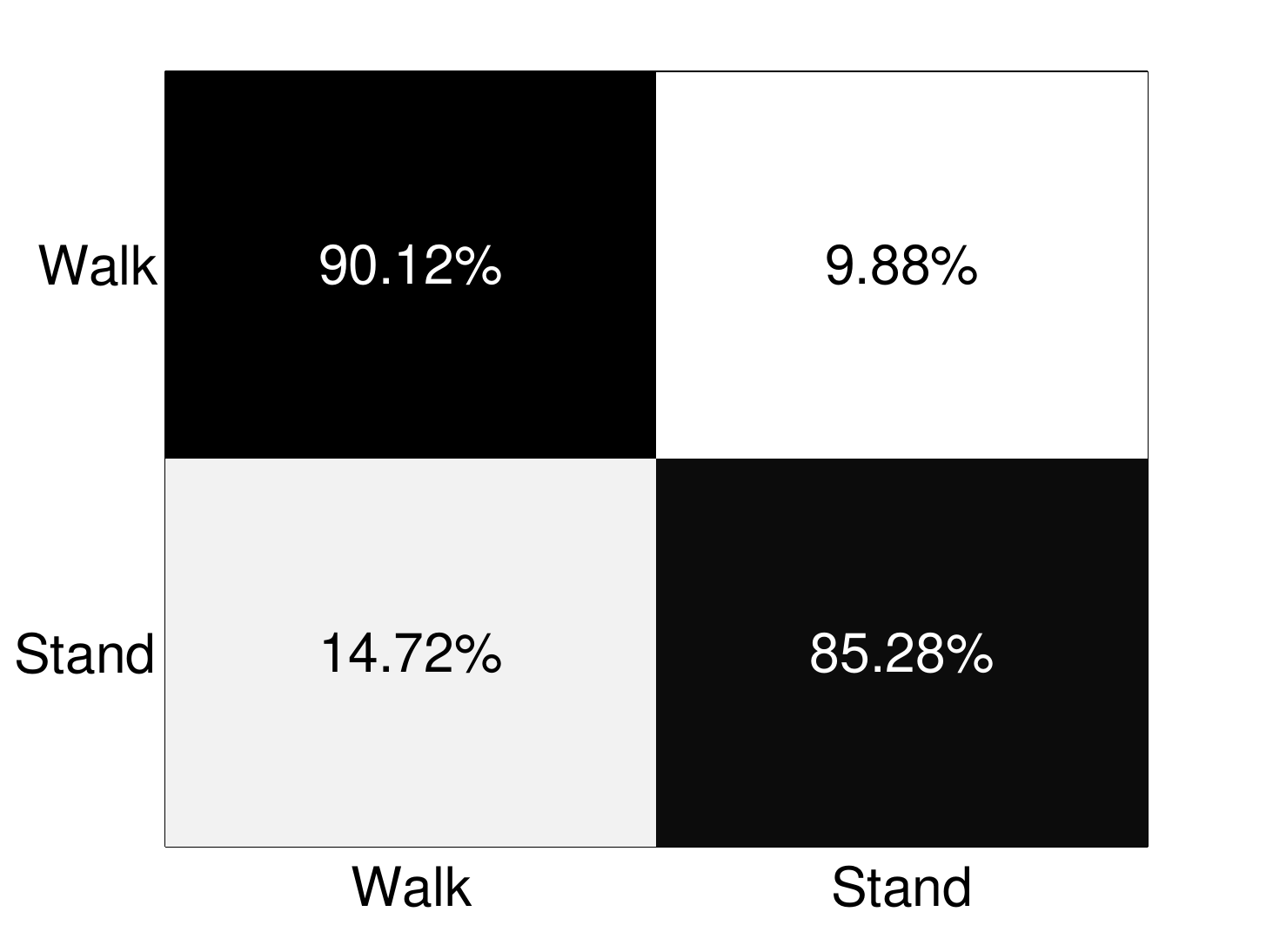}}
\caption{(a) Confusion matrix for collective activity $y$ from \cite{choi2012}. (b), (c), (d) Confusion matrices for collective activity, group activity and atomic action respectively form the proposed method. }
\label{fig:c_y}
\end{figure*}

We compare the collective activity results with \cite{eccv2014}, \cite{beyond_pami},\cite{choi2011} and \cite{choi2012} in the Table~\ref{table:act_compare}. To ensure a fair comparison with \cite{eccv2014} and \cite{beyond_pami}, we divide the dataset into separate training and testing sets as suggested by them. We use  leave-one-video-out method to compare with \cite{choi2011} as suggested. To compare with \cite{choi2012}, we use four fold setup with the splits mentioned by \cite{choi2012}. The Figure~\ref{fig:c_y} compares the confusion table of the proposed framework with that of \cite{choi2012}.  To find the accuracy for the group activity, we first identify the correctly detected groups and estimate accuracy for group activity on these groups. The confusion tables for group activity and atomic action are also given in Figure \ref{fig:c_y}. 

\begin{table}[h!]
\small
\caption{Comparison of overall and mean accuracies}
\begin{center}
\begin{tabular}{|p{10mm}|p{5mm}|p{5mm}|p{5mm}|p{5mm}|p{5mm}|p{5mm}|p{5mm}|}
\hline
Accuracy   & \multicolumn{3}{l|}{\cite{eccv2014} $/$ \cite{beyond_pami} $/$ Ours} & \multicolumn{2}{l|}{\cite{choi2011} $/$ Ours} & \multicolumn{2}{l|}{ \cite{choi2012} $/$ Ours} \\ \hline
Overall & -  &79.7  &81.1  &  -  &74.4  &79.1  & 76.3 \\ \hline
Mean Class & 92.0& 78.4 & 80.5 & 70.9 & 75.7 & 79.9 & 76.2  \\ \hline
\end{tabular}
\label{table:act_compare}
\end{center}
\end{table}

\begin{figure*}
\small
\centering
\subfloat[Two crossing groups]{\includegraphics[scale=0.13]{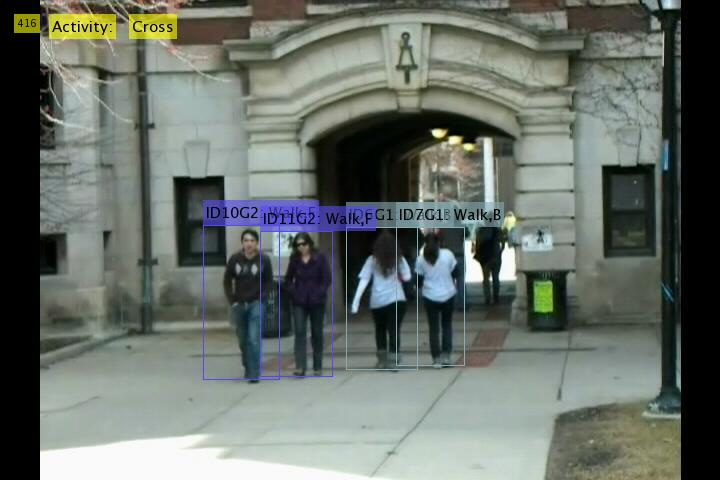}}
\subfloat[Two waiting groups]{\includegraphics[scale=0.13]{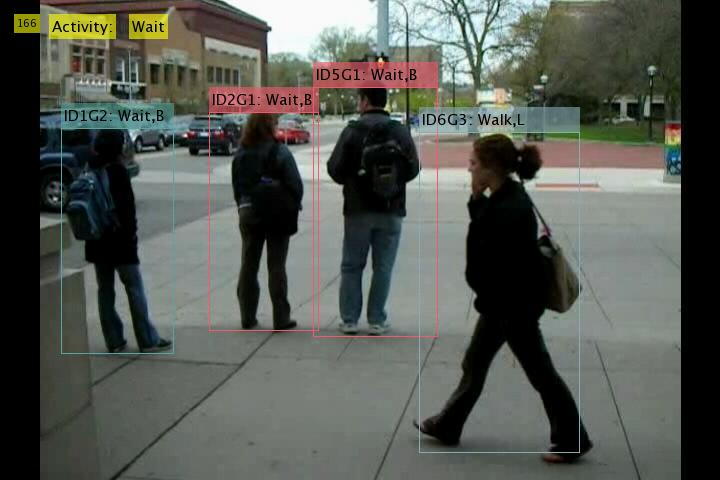}}
\subfloat[A group in a queue]{\includegraphics[scale=0.13]{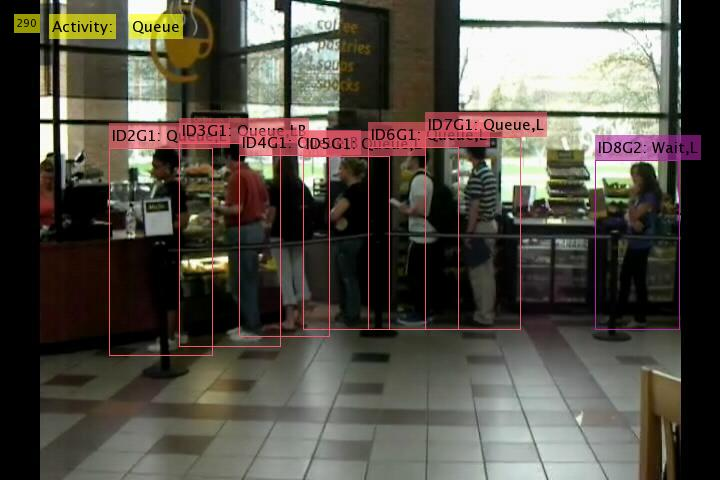}}
\subfloat[Two talking groups]{\includegraphics[scale=0.13]{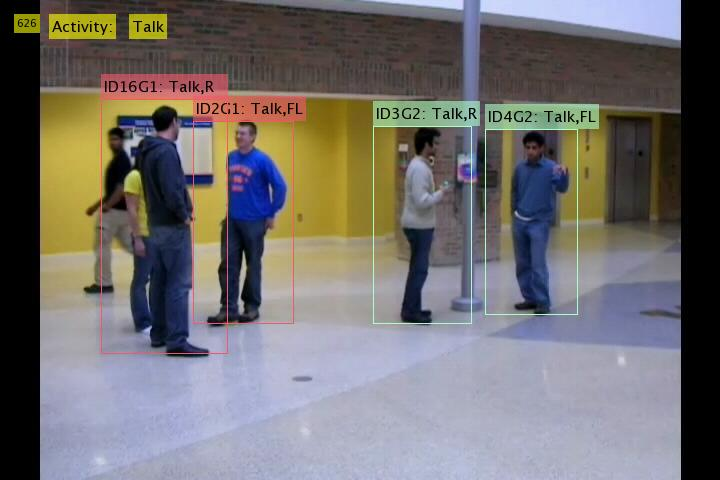}}
\subfloat[Two walking groups]{\includegraphics[scale=0.13]{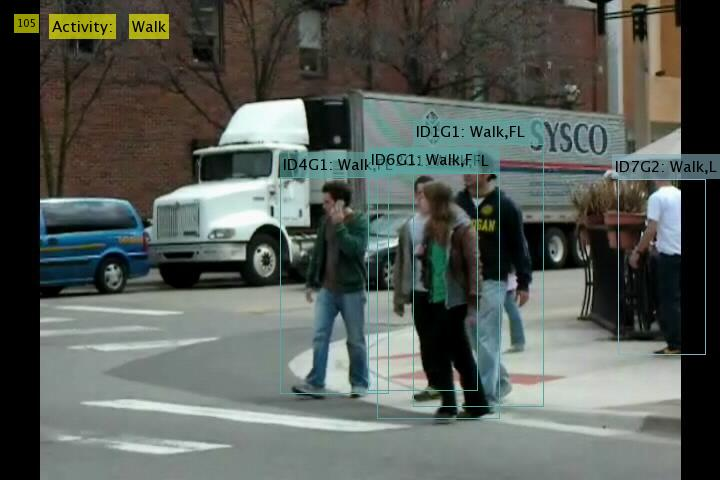}}\\
\subfloat[Two crossing groups]{\includegraphics[scale=0.13]{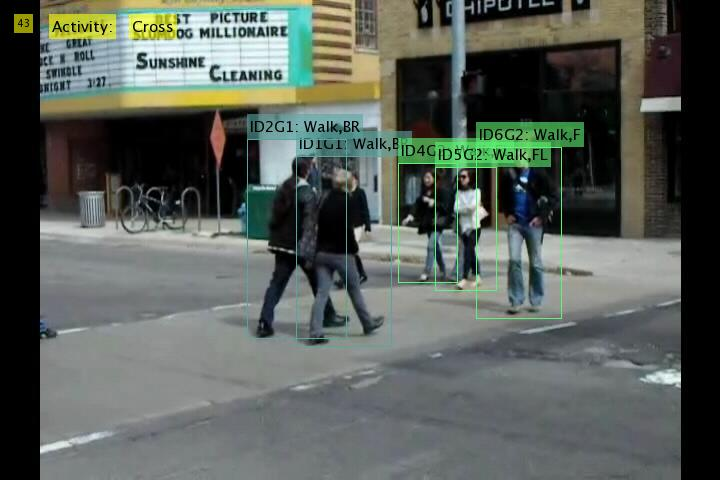}}
\subfloat[A waiting group]{\includegraphics[scale=0.13]{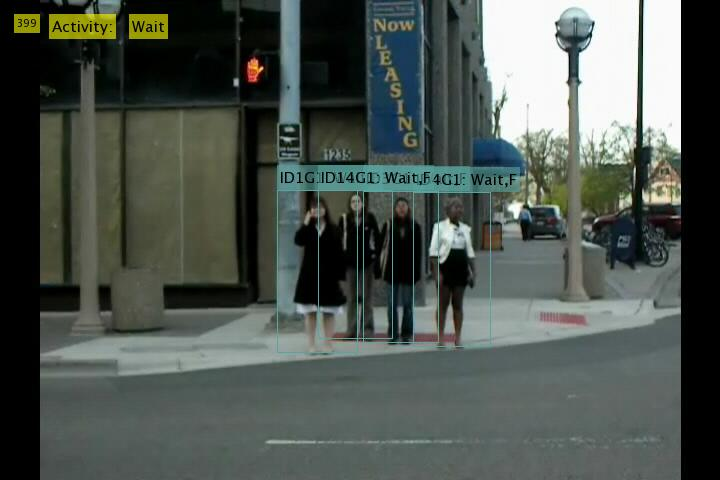}}
\subfloat[A group in a queue]{\includegraphics[scale=0.13]{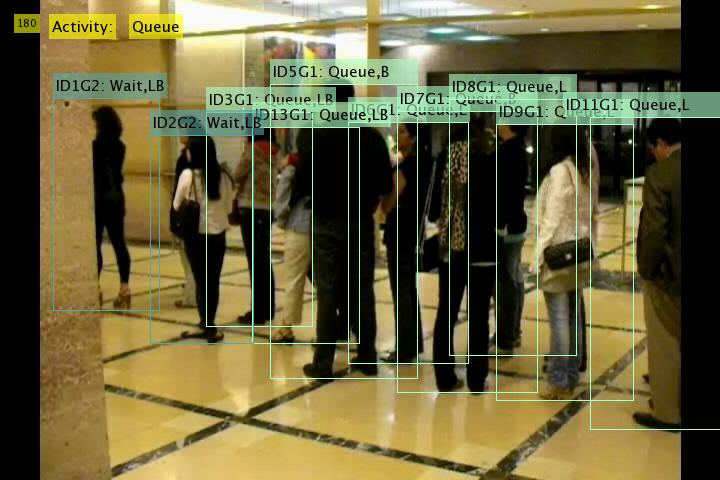}}
\subfloat[A talking group]{\includegraphics[scale=0.13]{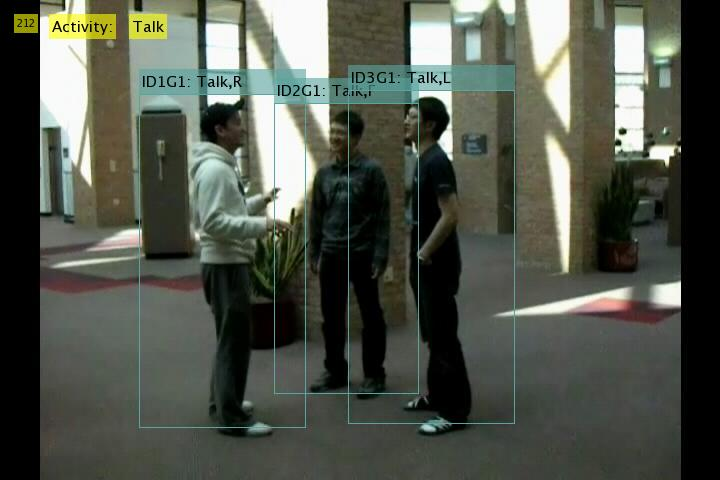}}
\subfloat[Three walking groups]{\includegraphics[scale=0.13]{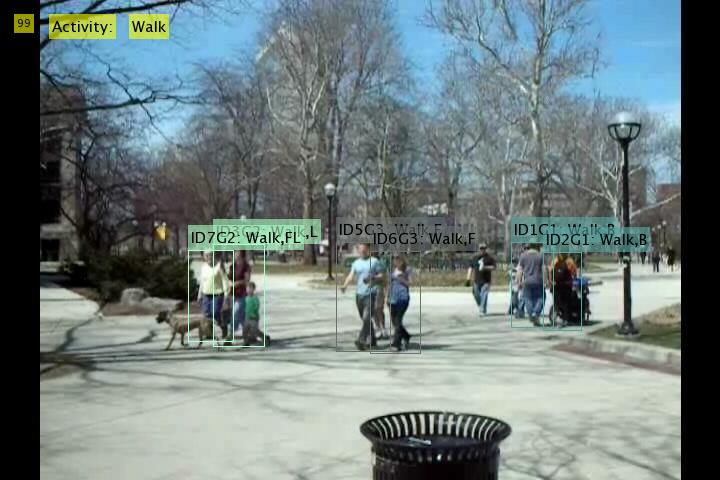}}
\caption{Qualitative results showing various collective and group activities. Collective activities column-wise: '\textit{cross}', '\textit{wait}', '\textit{queue}', '\textit{talk}', and '\textit{walk}'. A group is represented by a same color. Best viewed in color and when zoomed.}
\label{fig:out_img}
\end{figure*}

Form the Table~\ref{table:act_compare}, we notice that the proposed method offers a better accuracy than the methods \cite{beyond_pami} and \cite{choi2011}, and is marginally inferior to \cite{choi2012}. However, all these methods assume availability of either tracklets or \textit{action} labels whereas our method only needs the detections. Further, all these methods are non-causal in nature and involve batch processing of data unlike the proposed method. Additionally, we provide results at all levels of granularity (individual, group and collective). Figure \ref{fig:out_img} shows some qualitative results for group detection, group activity and collective activity. The members forming a group are represented by the same color. For example, Figure \ref{fig:out_img}(a) has two groups which are correctly identified as crossing each other. Hence the group activity for both the groups is \textit{walking} while the collective activity is \textit{crossing}. 

\subsection{Computational Performance}
Towards our main aim of developing a real-time system capable of simultaneous tracking, group detection and multi-level activity recognition, currently we achieve around 3 fps with our unoptimized MATLAB code on a i$7$ machine with $3.50$ GHz processor. With a proper implementation in GPU, we expect the frame rate to go up to 25 fps. To compare the computation time with one of the state-of-the-art algorithms, the method proposed in \cite{eccv2014} takes 6 hours of training and 120 s per inference whereas our proposed method takes around 90 s and 0.3 s respectively.

\section{Conclusions}
\label{section:con}
In this paper, we have proposed a novel approach for video understanding at various levels of granularity. We have presented a linear programming based method for joint estimation of tracks and groups. We have also proposed a method to recognize activities at various levels - individual, group and collective. The framework being causal in nature and computationally efficient, it is amenable for real-time implementation in video surveillance applications. The experiments show that the proposed method is very competitive with the state-of-the-art algorithms. 
{\small
\bibliographystyle{ieee}
\bibliography{Paper}
}
\end{document}